# Classifying traffic scenes using the GIST image descriptor

Ivan Sikirić
Mireo d.d.
Buzinski prilaz 32, 10000 Zagreb
e-mail: `ivan.sikiric@mireo.hr`

Karla Brkić, Siniša Šegvić
University of Zagreb
Faculty of Electrical Engineering and Computing
e-mail: `name.surname@fer.hr`

*Abstract*—This paper investigates classification of traffic scenes in a very low bandwidth scenario, where an image should be coded by a small number of features. We introduce a novel dataset, called the FM1 dataset, consisting of 5615 images of eight different traffic scenes: open highway, open road, settlement, tunnel, tunnel exit, toll booth, heavy traffic and the overpass. We evaluate the suitability of the GIST descriptor as a representation of these images, first by exploring the descriptor space using PCA and k-means clustering, and then by using an SVM classifier and recording its 10-fold cross-validation performance on the introduced FM1 dataset. The obtained recognition rates are very encouraging, indicating that the use of the GIST descriptor alone could be sufficiently descriptive even when very high performance is required.

## I. INTRODUCTION

This paper aims to improve current fleet management systems by introducing visual data obtained by cameras mounted inside vehicles. Fleet management systems track the position and monitor the status of many vehicles in a fleet, to improve the safety of their cargo and to prevent their unauthorized and improper use. They also accumulate this data for the purpose of generating various reports which are then used to optimise the use of resources and minimise the expenses. Fleet management systems usually consist of a central server to which hundreds of simple clients report their status. Clients are typically inexpensive embedded systems placed inside a vehicle, equipped with a GPS chip, a GPRS modem and optional additional sensors. Their purpose is to inform the server of the vehicle's current position, speed, fuel level, and other relevant information in regular intervals. Our contribution is introduction of visual cues to the vehicle's status. The server could use these cues to infer the properties of the vehicle's surroundings, which would help it in the further decision making. For example, server could infer the location of the vehicle (e.g. open road, tunnel, gas station), or cause of stopping (e.g. congestion, traffic lights, road works). In cases of missing or inaccurate GPS data, the fleet management server could use the location inferred from visual data to determine which of the several possible vehicle routes is the correct one. In some cases the server is not even aware of the loss of GPS precision, and fails to discard improbable data. This usually occurs in closed environments and near tall objects. Detecting such scenarios using visual data would be very beneficial. Also, detecting the cause of losing the GPS signal and the cause of vehicle stopping is important in systems that offer real-time tracking of vehicles used in transporting valuables.

We aim to solve this problem by learning a classification model for each scene or scenario.

Transmitting the entire image taken from a camera in every status report would raise the size of transmitted data by several orders of magnitude (typically the size of status is less than a hundred bytes), which would be too expensive. This can be resolved by calculating the descriptor of the image on the client itself, before transmitting it to the server for further analysis. We chose to use the GIST descriptor by Oliva and Torralba [1] because it describes the shape of the scene using low dimensional feature vectors, and it performs well in scene classification setups.

Our second contribution is introduction of a new dataset, called the FM1 dataset, containing 5615 traffic scenes and associated labels. Using this dataset we perform experimental evaluation of the proposed method and report some preliminary results. The remainder of the paper is organized as follows: In the next section we give an overview of previous related work, followed by a brief description of the GIST descriptor. We then describe the introduced dataset in detail, and explore the data using well known techniques. After that we define the classification problem, describe the classification setup and present the results. We conclude the paper by giving an overview of contributions and discussing some interesting future directions.

## II. RELATED WORK

Although image/scene classification is a very active topic in computer vision research, the work specific to road scene classification is limited. Bosch et al. [2] divide image classification approaches into low-level approaches and semantic approaches. Low-level approaches, e.g. [3], [4], model the image using low-level features, either globally or by dividing the image into sub-blocks. Semantic approaches aim to add a level of understanding *what* is in the image. Bosch et al. [2] identify three subtypes of semantic approaches: (i) methods based on semantic objects, e.g. [5], [6], where objects in the image are detected to aid scene classification, (ii) methods based on local semantic concepts, such as the bag-of-visual-words approach [7], [8], [9], where meaningful features are learned from the training data, and methods based on semantic properties, such as the GIST descriptor [1], [10], where an image is described by a set of statistical properties, such as roughness or naturalness. One might argue that the GIST descriptor should be categorized as a low-level approach, as it uses low-level features to obtain the representation. However, unlike low-level approaches [3], [4] where local features *are* the representation, the GIST descriptor merely *uses* low-level features to quantify higher-level semantic properties of the scene.

Ess et al. [11] propose a segmentation-based method for urban traffic scene understanding. An image is first divided into patches and roughly segmented, assigning each patch one of the thirteen object labels including car, road etc. This representation is used to construct a feature set fed to a classifier that distinguishes between different road scene categories.

Tang and Breckon [12] propose extracting a set of color, edge and texture-based features from predefined regions of interest within road scene images. There are three predefined regions of interest: (i) a rectangle near the center of the image, sampling the road surface, (ii) a tall rectangle on the left side of the image, sampling the road side, and (iii) a wide rectangle on the bottom of the image, sampling the road edge. Each predefined region of interest has its own set of preselected color features, including various components of RGB, HSV and YCrCb color spaces. The texture features are based on grey-level co-occurrence matrix statistics and Gabor filters. Additionally, edge-based features are extracted in the road edge region. A training set of 800 examples of four road scene categories is introduced: motorway, offroad, trunkroad and urban road. The testing is performed on approximately 600 test image frames. The k-NN and the artificial neural network classifiers are considered. The best obtained recognition rate is 86% when considering all four classes as separate categories, improving to 90% when classes are merged into two categories: off-road and urban.

Mioulet et al. [13] consider using Gabor features for road scene classification, using the dataset of Tang and Breckon [12]. Gabor features are extracted within the same three regions of interest used by Tang and Breckon. Grayscale histograms are built from Gabor result images and concatenated over all three regions of interest to form the final descriptor. A random forest classifier is trained and evaluated on a dataset from [12] consisting of four image classes: motorway, offroad, trunkroad and urban road, achieving the 10-fold cross-validation recognition rate of 97.6%.

In this paper, we are limited by our target application that imposes a constraint on bandwidth and processing time. Therefore, sophisticated approaches that require a lot of processing power to obtain the scene feature vector, such as the segmentation-based method of Ess et al. [11], or the many-feature method of Tang and Breckon [12], are unsuitable for our problem. The closest to our application is the work of Mioulet et al. [13], where a simple Gabor feature-based approach performs better on the same dataset than the various preselected features of Tang and Brackon. We take the idea of Mioulet et al. a step further by using Gabor feature-based GIST descriptor, which we hope will be an improvement over using raw Gabor features. Furthermore, as our goal is classifying the road scene into a much larger number of classes than are available in the dataset of Tang and Breckon [12], we introduce a new dataset of 8 scene categories.

III. THE GIST DESCRIPTOR

The GIST descriptor [1], [14] focuses on the shape of scene itself, on the relationship between the outlines of the surfaces and their properties, and ignores the local objects in the scene and their relationships. The representation of the structure of the scene, termed *spatial envelope* is defined, as well as its five perceptual properties: naturalness, openness, roughness, expansion and ruggedness, which are meaningful to human observers. The degrees of those properties can be examined using various techniques, such as Fourier transform and PCA. The contribution of spectral components at different spatial locations to spatial envelope properties is described with a function called windowed discriminant spectral template (WDST), and its parameters are obtained during learning phase.

The implementation we used first preprocesses the input image by converting it to grayscale, normalizing the intensities and locally scaling the contrast. The resulting image is then split into a grid on several scales, and the response of each cell is computed using a series of Gabor filters. All of the cell responses are concatenated to form the feature vector.

We believe this descriptor will perform very well in the context of traffic scenes classification. Expressing the degree of naturalness of the scene should enable us to differentiate urban from open road environments (even more so in case of unpaved roads). The degree of openness should differentiate the open roads from tunnels and other types of closed environments, and is especially useful in the context of fleet management, where closed environments usually result in the loss of GPS signal. Texture and shape of large surfaces is also taken into account, and this could help separate the highways from other types of roads.

IV. THE FM1 DATASET

Since we want to build a classifying model for traffic scenes and scenarios relevant to the fleet management problems, we introduce a novel dataset of labeled traffic scenes, called the FM1 dataset[1].

The data was acquired using a camera mounted on a windshield of a personal vehicle. Videos were recorded at 30 frames per second, with the resolution of 640x480. Several drives were made on Croatian roads, ranging in length from 37 to 60 minutes, to a total of about 190 minutes, see Table I. All videos were recorded on a clear sunny day, and the largest percentage of the footage was recorded on highways. The camera, its position and orientation were not changed between videos. We plan to vary the camera and its position in the later versions of the dataset, as well as to include nighttime images, and images taken during moderate to heavy rain (when the windshield wipers are operating).

A typical frame of a video is shown in Figure 1. Some parts of the image are not parts of the traffic scene, but rather the interior of a vehicle, such as camera mount visible in the upper right corner, the dashboard in the bottom part, and occasional reflections of car interior visible on the windshield. The windshield itself can be dirty, and various artefacts can appear on it depending on the position of the sun. The camera mount and the dashboard are in identical position in all of the images, so they should not influence the classification results. However, we can not expect this will always be the case in the future and plan to develop a system that will be robust enough to accommodate for those changes.

---

[1] http://www.zemris.fer.hr/~ssegvic/datasets/fer-fm1-2013-09.zip

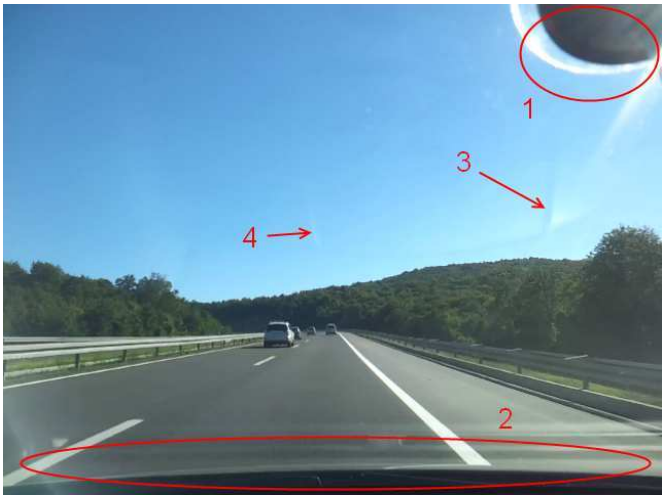

Fig. 1: A typical traffic scene. Note the camera mount (1), the dashboard (2), the reflection of car interior (3) and the speck of dirt (4)

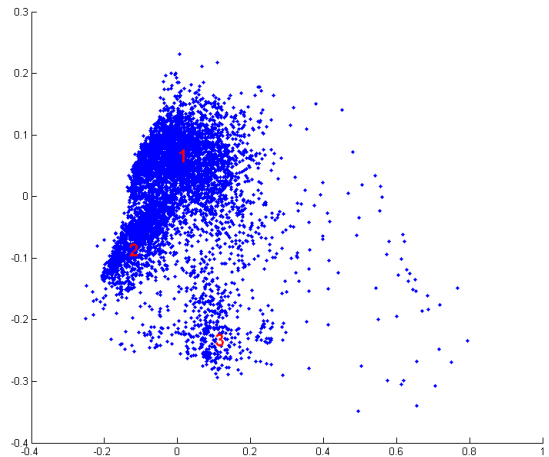

Fig. 2: Data points projected into 2D (note the three clusters)

TABLE I: Overview of traffic videos

| Video # | Duration | Number of extracted images |
|---|---|---|
| 1 | 54:08 | 1612 |
| 2 | 39:33 | 1181 |
| 3 | 1:00:00 | 1766 |
| 4 | 37:11 | 1106 |

For each recorded video, every 60th frame was extracted, i.e. one every two seconds. This ensured there was no bias in the selection process, and that subsequently selected images would not be too similar.

Choosing the classes for this preliminary classification evaluation was not an easy task. Perhaps the most obvious classes are "tunnel" and "open highway", but even they are not clearly defined. Should we insist the image be classified as an open highway if there is a traffic jam, and almost entire scene is obstructed by a large truck directly in front of us? We can introduce the "heavy traffic" class to accommodate for this case. What if it is not a truck, but a car instead, or if it is a bit further away, so more of the highway is visible? There are many cases in which it is impossible to set the exact moment when a scene transforms from being a member of one class and becomes a member of another class. Perhaps these issues can be resolved by allowing each scene to be assigned multiple class labels, which we plan to explore at later stages of our research.

Before choosing the class labels, we first analyzed the data using PCA and clustering algorithms, to see if there are obvious and easily separable classes in the dataset.

*A. Exploring the data*

The first step in data analysis was computing the descriptor for every image. We used the MATLAB implementation of the GIST descriptor provided by Oliva and Torralba [1], using the default parameters, which produces a feature vector of length 512. Since the resolution of the original image is 640x480, this provides a big dimensionality reduction.

The PCA was used to find the principal components in data, which was then projected into a plane determined by the first two principal components. The results can be seen in Figure 2. We can see most of the data points belong to one of three clusters, with some data points appearing to be outliers. Around 3300 points belong to the cluster 1, 90% of them corresponding to the scenes of open highway. The rest of the points in cluster 1 correspond to various types of scenes, but not a single one is a scene inside of a tunnel. Around 1400 data points belong to the cluster 2, 70% of them corresponding to the scenes of open highway, and 20% to the scenes of other types of open road. Around 820 points belong to the cluster 3, 45% of them corresponding to scenes inside the tunnels, 35% to the open highway, 6% of them to the scenes at the or under the toll booth. More than 95% of tunnel scenes belong to this cluster, and most of the images in this cluster depict closed environments.

The next step was using clustering algorithms to discover easily solvable classes in the dataset. K-means clustering was used on the dataset with values for K varying from 2 to 20. For the case K = 2, one cluster contained almost all of the tunnels and toll booth scenes, as was expected, but it also contained about one third of the open road scenes. For the case K = 3, one cluster contained most of the tunnels and toll booth scenes, but very few open road scenes. The second cluster contained almost all of the scenes with large number of other vehicles, i.e. scenes of heavy traffic. For K = 5, we can notice a good separation of non-highway open roads scenes. Further raising of K shows that similar types of scenes in the urban environments often end up in the same cluster. Some clusters contained very similar images, which suggested some easily separable classes, such as "a scene on a highway with a rocky formation on the right side of the road", but most of those classes were not considered useful to our application. In conclusion, the clustering approach gave us some ideas about easily separable and useful class labels to choose for our final classification experiment. Once we obtain more data, the clustering approach should be reapplied, probably in the form of hierarchical clustering for easier analysis.

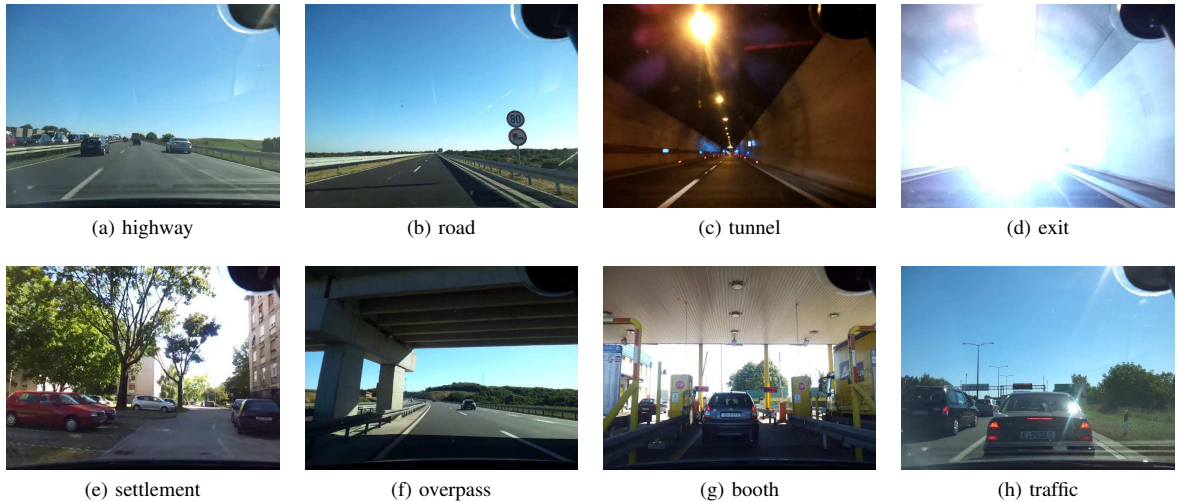

(a) highway  (b) road  (c) tunnel  (d) exit

(e) settlement  (f) overpass  (g) booth  (h) traffic

Fig. 3: Examples of selected classes

TABLE II: Selected classes

| class label | scene description |
|---|---|
| highway | an open highway |
| road | an open non-highway road |
| tunnel | in a tunnel, or directly in front of it, but not at the tunnel exit |
| exit | directly at the tunnel exit (extremely bright image) |
| settlement | in a settlement (e.g. visible buildings) |
| overpass | in front of, or under an overpass (the overpass is dominant in the scene) |
| booth | directly in front of, or at the toll booth |
| traffic | many vehicles are visible in the scene, or completely obstruct the view |

TABLE III: distribution of classes across videos

| Video # | highway | road | tunnel | exit | settlement | overpass | booth | traffic |
|---|---|---|---|---|---|---|---|---|
| 1 | 1382 | 0 | 185 | 12 | 0 | 8 | 0 | 0 |
| 2 | 652 | 312 | 134 | 7 | 59 | 3 | 3 | 0 |
| 3 | 1418 | 140 | 7 | 3 | 94 | 13 | 9 | 74 |
| 4 | 885 | 64 | 62 | 9 | 23 | 9 | 43 | 5 |
| Total | 4337 | 516 | 388 | 31 | 176 | 33 | 55 | 79 |

*B. Selected classes*

The set of classes we chose is listed in the Table II. Some of the classes were chosen because the data analysis indicated they would be easy to classify, and one of the functions of fleet management is to simply archive any data that might be required for purposes yet unknown. As was discussed in the introduction, we are very interested in detecting the environments in which the loss of GPS signal precision is likely, or the vehicle is likely to stop or drive slowly. The tunnel is a class which is both easy to classify and often causes loss of GPS signal. The tunnel exit was separated into its own class because the camera reaction to the sudden increase in sunlight is very slow, which results in extremely bright images (similar problem is not encountered during tunnel entry). The settlement is an environment in which we are likely to encounter tall objects (which may or may not be visible in the scene), so the loss of GPS signal precision is more likely to occur than on an open road, but less likely then in a tunnel. The overpass is chosen because going under it can cause a slight loss of GPS precision. Toll booths are included because going through them can cause a loss of GPS precision, and also because the vehicle must always stop at a toll booth. Heavy traffic scenario is interesting because it can cause the vehicle to stop or drive very slowly, and because presence of many vehicles can obstruct the view of the camera enough that the proper classification of the location becomes impossible. Intersection is a class which would be very interesting to investigate, but unfortunately we did not have enough samples, and their variability was too great. The distribution of class instances across videos is shown in Table III.

V. EXPERIMENTS

We used the data mining tool Weka 3 [15] to train several types of general purpose classifiers, performing grid search optimisation of parameters for each of them. The best results were obtained using SVM classifier with soft margin $C = 512$ and RBF kernel with $\gamma = 0.125$. The testing method was stratified cross-validation with 10 folds, using entire dataset (5615 feature vectors of length 512). Since the dataset is greatly biased towards the highway class, we expect most other classes to be often confused with a highway.

TABLE IV: Detailed accuracy by class

| Class | TP Rate | FP Rate | Precision | Recall | F-Measure | ROC Area |
|---|---|---|---|---|---|---|
| highway | 0.994 | 0.065 | 0.981 | 0.994 | 0.987 | 0.968 |
| settlement | 0.864 | 0.003 | 0.899 | 0.864 | 0.881 | 0.965 |
| booth | 0.855 | 0 | 0.979 | 0.855 | 0.913 | 0.99 |
| tunnel | 0.979 | 0.001 | 0.992 | 0.979 | 0.986 | 0.999 |
| exit | 0.903 | 0 | 0.933 | 0.903 | 0.918 | 1 |
| overpass | 0.485 | 0.001 | 0.696 | 0.485 | 0.571 | 0.95 |
| traffic | 0.861 | 0 | 0.971 | 0.861 | 0.913 | 0.986 |
| road | 0.899 | 0.007 | 0.93 | 0.899 | 0.914 | 0.957 |
| Weighted average | 0.973 | 0.051 | 0.973 | 0.973 | 0.973 | 0.969 |

TABLE V: Confusion matrix

| classified as: | highway | settlement | booth | tunnel | exit | overpass | traffic | road |
|---|---|---|---|---|---|---|---|---|
| highway | 4310 | 1 | 0 | 0 | 1 | 6 | 1 | 18 |
| settlement | 12 | 152 | 0 | 0 | 0 | 0 | 0 | 12 |
| booth | 5 | 2 | 47 | 0 | 0 | 0 | 0 | 1 |
| tunnel | 6 | 1 | 0 | 380 | 1 | 0 | 0 | 0 |
| exit | 1 | 0 | 0 | 2 | 28 | 0 | 0 | 0 |
| overpass | 15 | 0 | 0 | 1 | 0 | 16 | 0 | 1 |
| traffic | 5 | 2 | 1 | 0 | 0 | 0 | 68 | 3 |
| road | 39 | 11 | 0 | 0 | 0 | 1 | 1 | 464 |

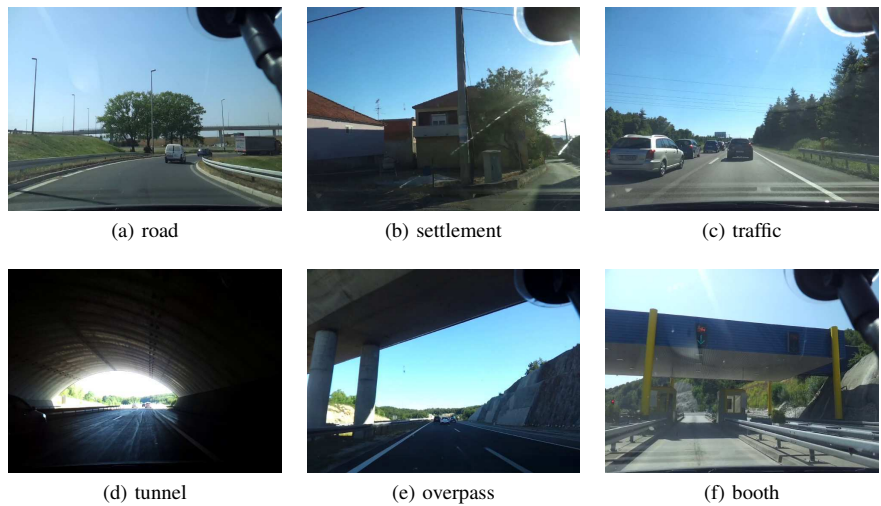

(a) road  (b) settlement  (c) traffic

(d) tunnel  (e) overpass  (f) booth

Fig. 4: Examples of scenes misclassified as highway

The achieved recognition rate was 97.3%. The detailed accuracy by class is shown in Table IV, while the confusion matrix is shown in Table V.

We can see the performance is very good across all classes, except the overpass, which is often confused with a highway. This is probably because most of the overpass scenes were in fact on a highway, and the overpass was not equally dominant in all of them, as well as because of low number of class instances. The settlement class is often confused with highway and road scenes. We plan to resolve this by introducing more data, because instances of this class vary in appearance more then of any other class. We also note that highway is often confused with a plain road, and vice versa, which is to be expected for classes with similar appearance. We do not consider this confusion to be problematic, as fleet management can usually use GPS data to correctly infer the type of the road. Some examples of images misclassified as a highway are shown in Figure 4.

## VI. CONCLUSION AND FUTURE WORK

Our preliminary results show that the GIST descriptor alone is sufficiently descriptive for the purpose of classification of many types of traffic scenes. This indicates viability of the proposed method as an improvement of current fleet management systems, and invites further research.

Further efforts should be directed towards expanding the traffic scenes dataset, to include images during nighttime, and bad weather, as well as images obtained by different cameras mounted at other positions and angles. Also, the set of classes should be expanded to include many other interesting scenes and scenarios. Poor classification of the overpass class should be addressed, at first by adding more data. It is possible that in some cases the overpass can not be considered an integral part of the scene, but rather its attribute. There are also other types of attributes of the traffic scenes that we might want to consider, e.g. traffic lights. We plan to introduce additional image features and detectors of such attributes.

One of our future goals would be to reduce or eliminate the need for labeling. We are considering using automatic labeling in cases where we can infer the details of vehicle's environment with high degree of probability using non-visual cues.

ACKNOWLEDGMENTS

The authors would like to thank Josip Krapac for his very helpful input during our work on this paper.

This research has been supported by the research projects Research Centre for Advanced Cooperative Systems (EU FP7 #285939) and Vista (EuropeAid/131920/M/ACT/HR).

REFERENCES

[1] A. Oliva and A. Torralba, "Modeling the shape of the scene: A holistic representation of the spatial envelope," *Int. J. Comput. Vision*, vol. 42, pp. 145–175, May 2001.

[2] A. Bosch, X. Muñoz, and R. Martí, "Review: Which is the best way to organize/classify images by content?," *Image Vision Comput.*, vol. 25, pp. 778–791, June 2007.

[3] N. Serrano, A. E. Savakis, and J. Luo, "Improved Scene Classification using Efficient Low-Level Features and Semantic Cues," *Pattern Recognition*, vol. 37, no. 9, pp. 1773–1784, 2004.

[4] A. Vailaya, M. A. T. Figueiredo, A. K. Jain, and H. J. Zhang, "Image classification for content-based indexing," *IEEE Trans. on Image Processing*, vol. 10, pp. 117–130, January 2001.

[5] J. Luo, A. E. Savakis, and A. Singhal, "A bayesian network-based framework for semantic image understanding," *Pattern Recogn.*, vol. 38, pp. 919–934, June 2005.

[6] J. Fan, Y. Gao, H. Luo, and G. Xu, "Statistical modeling and conceptualization of natural images," *Pattern Recogn.*, vol. 38, pp. 865–885, June 2005.

[7] F.-F. Li and P. Perona, "A bayesian hierarchical model for learning natural scene categories," in *Proceedings of the 2005 IEEE Computer Society Conference on Computer Vision and Pattern Recognition (CVPR'05) - Volume 2 - Volume 02*, CVPR '05, (Washington, DC, USA), pp. 524–531, IEEE Computer Society, 2005.

[8] A. Bosch, A. Zisserman, and X. Muñoz, "Scene classification via plsa," in *Proceedings of the 9th European conference on Computer Vision - Volume Part IV*, ECCV'06, (Berlin, Heidelberg), pp. 517–530, Springer-Verlag, 2006.

[9] S. Lazebnik, C. Schmid, and J. Ponce, "Beyond bags of features: Spatial pyramid matching for recognizing natural scene categories," in *Proceedings of the 2006 IEEE Computer Society Conference on Computer Vision and Pattern Recognition - Volume 2*, CVPR '06, (Washington, DC, USA), pp. 2169–2178, IEEE Computer Society, 2006.

[10] A. Oliva and A. B. Torralba, "Scene-centered description from spatial envelope properties," in *Proceedings of the Second International Workshop on Biologically Motivated Computer Vision*, BMCV '02, (London, UK, UK), pp. 263–272, Springer-Verlag, 2002.

[11] A. Ess, T. Mueller, H. Grabner, and L. v. Gool, "Segmentation-based urban traffic scene understanding," in *Proceedings of the British Machine Vision Conference*, pp. 84.1–84.11, BMVA Press, 2009. doi:10.5244/C.23.84.

[12] I. Tang and T. Breckon, "Automatic road environment classification," *IEEE Transactions on Intelligent Transportation Systems*, vol. 12, pp. 476–484, June 2011.

[13] L. Mioulet, T. Breckon, A. Mouton, H. Liang, and T. Morie, "Gabor features for real-time road environment classification," in *Proc. International Conference on Industrial Technology*, pp. 1117–1121, IEEE, February 2013.

[14] M. Douze, H. Jégou, H. Sandhawalia, L. Amsaleg, and C. Schmid, "Evaluation of gist descriptors for web-scale image search," in *Proceedings of the ACM International Conference on Image and Video Retrieval*, CIVR '09, (New York, NY, USA), pp. 19:1–19:8, ACM, 2009.

[15] M. Hall, E. Frank, G. Holmes, B. Pfahringer, P. Reutemann, and I. H. Witten, "The weka data mining software: an update," *SIGKDD Explor. Newsl.*, vol. 11, pp. 10–18, Nov. 2009.